\documentclass[final,3p,times]{elsarticle}

\usepackage{amssymb}

\usepackage[caption=false,font=footnotesize]{subfig}
\usepackage{algorithm}
\usepackage{algorithmic}
\usepackage{graphicx}
\usepackage[ansinew]{inputenc}

\journal{Parallel Computing}

\begin{document}

\begin{frontmatter}

\title{A parallel evolutionary algorithm to optimize dynamic memory managers in embedded systems}

\author[addr1]{Jos\'{e} L. Risco-Mart\'{i}n}
\author[addr2,addr1]{David Atienza}
\author[addr3]{J. Manuel Colmenar}
\author[addr1]{Oscar Garnica}
\address[addr1]{Department of Computer Architecture and Automation, Universidad Complutense de Madrid, 28040 Madrid, Spain}
\address[addr2]{Embedded Systems Laboratory (ESL), Ecole Polytechnique F\'ed\'erale de Lausanne (EPFL), 1015 Lausanne, Switzerland}
\address[addr3]{C. E. S. Felipe II, Universidad Complutense de Madrid, 28300 Aranjuez, Spain}

\begin{abstract}
For the last thirty years, several \emph{Dynamic Memory Managers (DMMs)} have been proposed. Such DMMs include first fit, best fit, segregated fit and buddy systems. Since the performance, memory usage and energy consumption of each DMM differs, software engineers often face difficult choices in selecting the most suitable approach for their applications. This issue has special impact in the field of portable consumer embedded systems, that must execute a limited amount of multimedia applications (e.g., 3D games, video players and signal processing software, etc.), demanding high performance and extensive memory usage at a low energy consumption. Recently, we have developed a novel methodology based on genetic programming to automatically design custom DMMs, optimizing performance, memory usage and energy consumption. However, although this process is automatic and faster than state-of-the-art optimizations, it demands intensive computation, resulting in a time consuming process. Thus, parallel processing can be very useful to enable to explore more solutions spending the same time, as well as to implement new algorithms. In this paper we present a novel parallel evolutionary algorithm for DMMs optimization in embedded systems, based on the \emph{Discrete Event Specification (DEVS)} formalism over a \emph{Service Oriented Architecture (SOA)} framework. Parallelism significantly improves the performance of the sequential exploration algorithm. On the one hand, when the number of generations are the same in both approaches, our parallel optimization framework is able to reach a speed-up of 86.40$\times$ when compared with other state-of-the-art approaches. On the other, it improves the global quality (i.e., level of performance, low memory usage and low energy consumption) of the final DMM obtained in a 36.36\% with respect to two well-known general-purpose DMMs and two state-of-the-art optimization methodologies.
\end{abstract}


\end{frontmatter}

\section{Introduction and related work}
\label{sec:Introduction}

Modern multimedia embedded systems must execute applications coming from desktop systems. As a result, one of the most important problems that system designers face today is the integration of a great amount of applications coming from a general-purpose domain into a highly-constrained memory device \cite{RiscoMartin2008a}. Usually, these applications are developed in the C++ language, where dynamic memory is allocated via the operator \texttt{new} and deallocated by the operator \texttt{delete}. Most compilers simply map these operators directly to the \texttt{malloc()} and \texttt{free()} functions of the standard C library. Studies have shown that dynamic memory management can consume up to 38\% of the execution time in C++ applications \cite{Calder1994}. Thus, the performance of dynamic memory management can have a substantial effect on the overall performance of C++ applications. In the past, most of the implementations that were ported to these embedded platforms stayed mainly in the classic domain of signal processing. Such implementations actively avoided algorithms that employ dynamic memory.

New portable devices must rely on dynamic memory for a very significant part of their functionality due to the inherent unpredictability of the input data. These devices also integrate multiple services such as multimedia and wireless network communications. It heavily influences global performance and memory usage of the system. In addition, energy consumption has become a real issue in overall system design (both embedded and general-purpose) due to circuit reliability and packaging costs \cite{Vijaykrishnan2003}. Thus, the optimization of the dynamic memory subsystem in general-purpose devices, and particularly in embedded systems, has three goals that cannot be seen independently: performance, memory usage and energy consumption.

Since the dynamic memory subsystem heavily influences performance and is an important source of energy consumption and memory usage, flexible system-level implementation and evaluation mechanisms for these three factors must be available at an early stage of the design flow for embedded systems. Current implementations of \emph{Dynamic Memory Managers (DMMs)} can provide a reasonable level of performance for general-purpose systems \cite{Wilson1995}. However, these implementations do not consider energy consumption and memory usage constraints of the target embedded platforms where these DMMs must run on. Thus, these general-purpose DMMs implementations are never optimal for the final target embedded platform and produce large energy and performance penalties. Consequently, system designers currently face the need to manually optimize the implementations of the initial DMMs on a case-per-case basis. However, adding new implementations of (complex) custom DMMs manually often prove to be a very programming-intensive and error-prone task that consumes a significant part of the time spent in system integration of dynamic memory management mechanisms, even if standardized languages such as C or C++ offer considerable support~\cite{Berger2001,Wilson1995}. 

Therefore, new DMM designs are needed to optimize the execution of embedded applications, where indeed each application presents an independent memory behavior and, as a consequence, particular DMM implementations need to be built for each one of them. In this regard, Kiem-Phong Vo presents a DMM that allows the definition of multiple memory regions with different disciplines \cite{Vo1996}. However, this approach cannot be extended with new functionality and is limited to a small set of user-defined functions for memory de/allocation.

Berger \emph{et al} propose an infrastructure of C++ layers that can be used to improve performance of general-purpose DMMs \cite{Berger2001}. The problem with these implementations is that they mainly aim at performance optimizations and propose ad-hoc solutions, not being possible to automatically explore and implement optimal DMMs in a methodical way for different embedded systems, as we propose in this work. However, Berger \emph{et al} state that the versatility of the C++ language allows to insert this new principle of extensive modular library inside embedded operating systems, such as RTEMS \cite{RTEMS}, without the code size overhead found in other approaches that optimize DMMs in C code based on profiling \cite{Attardi1999}.

Closer to our work, in \cite{Atienza2006} and \cite{Atienza2006a}, a methodology is defined to implement a DMM by means of a pseudo-random exploration, which is realized in a complete design space for dynamic embedded systems. The authors aim to reduce energy consumption \cite{Atienza2006} and memory usage \cite{Atienza2006a}. In addition, this study may be particularized for each embedded application. Using the same methodology, in \cite{Mamagkakis2006} the authors balance two factors (memory usage and memory accesses) in order to achieve the most energy efficiency. However, they need two preliminary phases that require at least two human decisions: (1) the significant reduction of the initial design space, and (2) the execution of every DMM to evaluate its behavior, which requires to compile and run the target application in each case.

Recently, we have proposed a new method, using genetic programming, to automatically generate optimal DMM implementations, thus improving the state-of-the-art exploration approaches \cite{RiscoMartin2009b}. We have introduced a novel approach that allows developers to design custom DMMs with the optimized performance, memory usage and energy consumption required for these new dynamic multimedia applications, with no manual intervention in the exploration effort. To this end, we have used grammatical evolution \cite{ONeill2003} to traverse the design space according to the dynamic memory behavior of these new multimedia applications. To evaluate each DMM implementation found by our evolutionary algorithm, we have implemented a DMM simulator , extending the DMM library developed in \cite{Atienza2006}. It enables a fast evaluation of each DMM implementation for fine-tuning our dynamic memory design space exploration results during the optimization process. However, although the proposed methodology is up to 20$\times$ faster than previous approaches, the optimization time (16 hours in the worst case for large embedded applications) can be considerably improved.

In this work, we propose a new parallel optimization methodology, which combines genetic programming with a parallel optimization scheme, designed to automatically generate optimal DMM implementations, thus improving the state-of-the-art exploration approaches. Existing parallel implementations of evolutionary algorithms can be classified into three main types \cite{CantuPaz1998}: (1) global single-population master-worker algorithms, (2) single-population fine-grained, and (3) multiple-population coarse-grained algorithms. Taking into account this classification, our parallel design may be included in the first group. The proposed parallel algorithm is formally implemented using \emph{Discrete Event Specification over Service Oriented Architecture (DEVS/SOA)} \cite{Mittal2009a}, which we have adapted to use multiple distributed processing cores over local networks and web services (for massive parallelization).

Our approach allows developers to design custom DMMs with the reduced execution time, memory usage and power consumption required for the dynamic multimedia applications run on embedded systems, and with no manual intervention in the exploration effort. First, starting from all the possible DMM implementations that the aforementioned methodology proposes (\cite{RiscoMartin2009b}), we automatically define the relevant design space of dynamic memory management decisions for an optimal performance, memory usage and energy consumption. After that, using grammatical evolution and a parallel scenario, we traverse this design space according to the DMM behavior of the dynamic applications. To evaluate each DMM implementation found by our evolutionary algorithm, we have developed a DMM simulator. It enables a relatively fast evaluation of each DMM implementation for fine-tuning our DMM design space exploration results during the optimization process.

Overall, our experiments in two real-life dynamic embedded applications show that our novel parallel DMM exploration approach, using grammatical evolution and parallel optimization, achieves important speed-ups (up to 66.2 times faster) with respect to other two state-of-the-art approaches. Moreover, the proposed parallel exploration framework can achieve a 50.25\% better set of trade-off solutions (performance, power consumption and area usage) of DMMs implementations; thus, it can generate a much better approximation to the Pareto-optimal front of DMMs solutions for designers to choose from.

The remainder of the article is organized as follows. First, Section \ref{sec:TheProblem} describes the dynamic memory manager design space and how grammatical evolution is applied to generate DMMs. Then, in Section \ref{sec:OptimizationFlow}, we present our design flow to automatically explore DMMs, optimizing performance, memory usage and energy consumption. A description of our parallel proposal, which uses a master-worker scheme is detailed in Section \ref{sec:ParallelImplementation}. In Section \ref{sec:Experiments}, we describe our experimental setup for parallel simulations and we discuss the obtained results in two real-life multimedia applications for embedded systems. Finally, Section \ref{sec:Conclusions} summarizes the main conclusions of this paper as well as our future work.


\section{The dynamic memory managers exploration problem}
\label{sec:TheProblem}

In this section we first summarize the main characteristics of a DMM. Next, we present the taxonomy designed to classify the DMMs design space, and briefly describe the C++ template library that implements the given taxonomy (in both simulation and real mode). Finally, we describe how grammatical evolution is applied to automatically explore the design space.

\subsection{Introduction to dynamic memory managers}

Dynamic memory management basically consists of two separate tasks, i.e., allocation and deallocation. Allocation is the mechanism that searches for a memory block big enough to satisfy the memory requirements of an object request in a given application. Deallocation is the mechanism that returns a freed memory block to the available memory of the system in order to be reused subsequently. In current applications, the blocks are requested and returned in any order. The amount of memory used by the DMM grows up when the memory storage space is used inefficiently, reducing the storage capacity. This phenomenon is called fragmentation. Internal fragmentation happens when requested objects are allocated in blocks whose size is bigger than the size of the object. External fragmentation occurs when no blocks are found for a given object request despite enough free memory is available. Hence, on top of memory de/allocation, the DMM has to take care of memory usage issues. To avoid these problems, some DMMs include splitting (breaking large blocks into smaller ones to allocate a larger number of small objects) and coalescing (combining small blocks into bigger ones to allocate objects for which there are no available blocks of their size). However, these two algorithms usually reduce performance, and consume more energy as well. To support these mechanisms, additional data structures are built to keep track of the free and used blocks.

There are several general-purpose DMMs. Here we briefly describe two of them: the Kingsley DMM \cite{Wilson1995} and the Lea DMM \cite{Lea2009}. Our optimized DMMs are compared with them because Kingsley and Lea are widely used in both general-purpose and embedded systems. Moreover, they are on opposite ends between maximizing performance and minimizing memory usage. The Kingsley DMM was originally used in BSD 4.2. Windows-based systems (both mobile and desktop) apply the main ideas from Kingsley. The Kingsley DMM organizes the available memory in power-of-two block sizes: all allocation requests are rounded up to the next power of two. This rounding can lead to severe memory usage issues because, in the worst case, it allocates twice as much memory as requested. It performs no splitting or coalescing. This algorithm is well known to be among the fastest DMMs although it is among the worst in terms of memory usage. On the contrary, the Lea DMM is an approximate best-fit DMM that provides mainly low memory usage. Linux-based systems implement as their basis the Lea DMM. It presents a different behavior based on the size of memory requested. For example, small objects (less than 64 bytes) are allocated in free blocks using an exact-fit policy (one linked list of blocks for each multiple of 8 bytes). For medium-sized objects (less than 128K), the Lea allocator performs immediate coalescing and splitting in the previous lists and approximates best-fit. For large objects ($\geq$ 128K), it uses virtual memory (through the \texttt{mmap} primitive).

To create an efficient DMM, the design decisions that can be taken to handle the possible combinations of the previous factors (e.g., performance, memory usage, energy consumption, overhead of additional data structures, etc) must be classified. Next, we describe the taxonomy defined to classify such design decisions.

\subsection{Dynamic memory management design space for embedded systems}

\begin{figure}[!t]
\centering
\includegraphics[width=0.60\columnwidth]{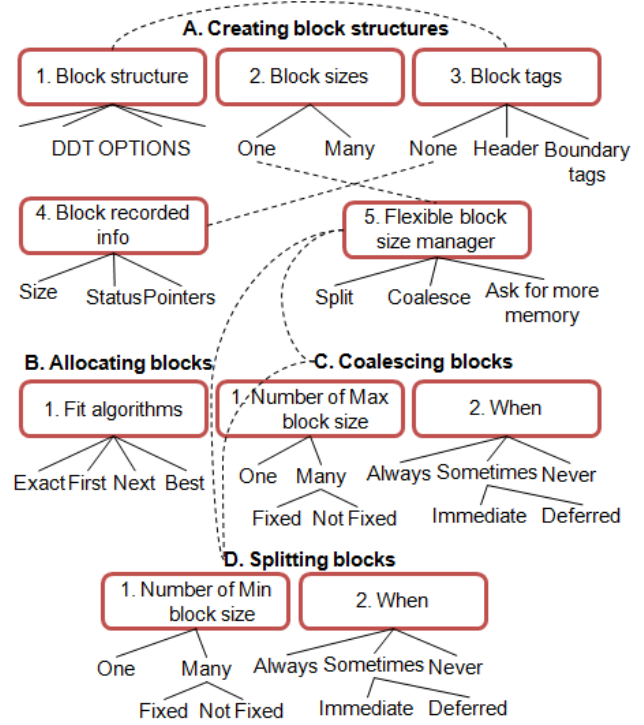}
\caption{Dynamic memory management search space of corthogonal decisions.}
\label{fig:OrthogonalTrees}
\end{figure}

A DMM is formed by one or more \emph{Atomic Dynamic Memory Managers (ADMs)}. An ADM is defined for a specific behavior pattern. The definition of an ADM can be classified in five different categories~cite{Atienza2006a}, which are depicted in Fig. \ref{fig:OrthogonalTrees} and described below:

\begin{enumerate}[A.]

\item \emph{Creating block structures}. This category handles the way block data structures are created and later used by the ADM to satisfy the memory requests. More specifically, the \textit{Block structure} tree in Fig. \ref{fig:OrthogonalTrees} specifies the data structure that manages the free blocks (singly-linked list, doubly-linked list, AVL trees, etc). The \emph{Block sizes} tree refers to the different sizes of basic blocks available in the ADM, which may be fixed or not. The \emph{Block tags} and the \emph{Block recorded info} trees specify the extra fields needed inside the block to store information used by the ADM. Finally, the \emph{Flexible block size manager} tree decides if the splitting and coalescing mechanisms are activated according to the availability of the size of the requested memory block.

\item \emph{Allocating blocks}. It deals with the allocation policy applied to the ADM. Here we may include all the important choices available in order to choose a block from a list of free blocks \cite{Wilson1995}.

\item \emph{Coalescing blocks}. It concerns the actions executed by the ADM to merge two smaller blocks into a larger one. Firstly, the \emph{Number of max block size} tree defines the new block sizes that are allowed after coalescing two different adjacent blocks. Then, the \emph{When} tree defines how often coalescing should be performed.

\item \emph{Splitting blocks}. This category refers to the actions executed by the ADM to split one larger block into two smaller ones. Firstly, the \emph{Number of min block size} tree defines the new block sizes that are allowed after splitting a block into smaller ones. And the \emph{When} tree defines how often splitting should be performed.

\end{enumerate}

In summary, when one decision has been taken in every tree, one custom ADM is defined. However, the decision categories and trees presented above include some interdependencies (constraints, in our design space). Therefore, the selection of certain leaves in some trees heavily affects the coherent decisions in the others. These dependencies are represented using dashed lines in Fig. \ref{fig:OrthogonalTrees}. First, the \emph{Block structure} tree inside the \emph{Creating block structures} category strongly influences the decision in the \emph{Block tags} tree of the same category because certain data structures require extra fields for their maintenance. For example, singly-linked lists, where several block sizes are allowed, have to include a header field and the size of each free block inside. Second, inside the same category, if the \texttt{None} leaf from the \emph{Block tags} tree is selected, then the \emph{Block recorded info} tree cannot be used. Clearly, there would be no memory space to store the recorded info inside the block. In the same manner, the \texttt{One} leaf from the \emph{Block sizes} tree, excludes the use of the \emph{Flexible block size manager} tree in the \emph{Creating block structures} category. This occurs because the one block size leaf does not allow us to define any new block size. Similarly, the coalescing and splitting mechanisms are quite related and the decisions in one category have to find equivalent ones in the other. Finally, the \emph{Flexible block size manager} tree heavily influences all the trees inside the \emph{Coalescing Blocks} and the \emph{Splitting Blocks} categories.

This new approach allows us the reduction of the complexity of the DMM global design in smaller sub-problems. Table \ref{tab:Kingsley} shows an excerpt of the Kingsley DMM using our taxonomy. Each column represents one tree in Fig. \ref{fig:OrthogonalTrees}, labeled as category and tree number. Thus, the Kingsley DMM is formed by 30 ADMs. Regarding the \emph{Creating block structures} category, the \emph{Singly-Linked List (SLL)} is used in the \emph{Block structure} tree. As column A.2 shows, every ADM manages one block size (in powers of two). We select \texttt{Header} in the \emph{Boundary tags} tree because the SLL contains a next pointer field. Furthermore, the header also contains the size, which is a constant (column A.4 in Table \ref{tab:Kingsley}). The allocation policy employed by all the ADMs in Kingsley is \texttt{First fit}, shown in column B.1. Finally, columns A.5, C.1, C.2, D.1 and D.2 are not applicable to Kingsley, because it does not use splitting or coalescing algorithms. In the following, we describe how the previous taxonomy has been implemented.

\begin{table*}[ht]
\caption{Definition of the Kingsley DMM using our taxonomy. Every column represents a tree of Fig. \ref{fig:OrthogonalTrees}}
\begin{center}
\begin{tabular}{|c|c|c|c|c|c|c|c|c|c|}
\hline
\textbf{A.1} & \textbf{A.2} & \textbf{A.3} & \textbf{A.4} & \textbf{A.5} & \textbf{B.1} & \textbf{C.1} & \textbf{C.2} & \textbf{D.1} & \textbf{D.2} \\
\hline
SLL & One: $2^3$B & Header & Size \& Pointers & - & First & - & - & - & - \\
\hline
SLL & One: $2^4$B & Header & Size \& Pointers & - & First & - & - & - & - \\
\hline
SLL & One: $2^5$B & Header & Size \& Pointers & - & First & - & - & - & - \\
\hline
\multicolumn{10}{|c|}{$\ldots$} \\
\hline
SLL & One: $2^{32}$B & Header & Size \& Pointers & - & First & - & - & - & - \\
\hline
\end{tabular}
\end{center}
\label{tab:Kingsley}
\end{table*}

We have developed a C++ library based on abstract classes and templates \cite{Stroustrup2000}, with which we may implement all the possible decisions in the DMM design space depicted in Fig. \ref{fig:OrthogonalTrees}. To this end, we have followed the same structure developed by Berger \cite{Berger2001} and Atienza \emph{et al} \cite{Atienza2006}, developing several data structures and de/allocation policies, utility layers, object representations, selectors, headers, etc. This template-based approach largely simplifies the complex engineering process of designing custom DMMs, allowing the developers to cover a vast part of the implementation space (e.g., different strategies of the DMM, internal blocks of the allocators, etc.) with a minimal programming and modeling effort. In addition, this library provides a logging layer, that reports at runtime the number, type and size of objects de/allocated by each application under study.

Thus, our library enables the construction of the final global custom DMM implementation in a simple way via composition of C++ layers. In general terms, the basic interface defined in such DMM library, called \emph{AtomicDMM}, is based on a C++ template. As stated before, every DMM is formed by a set of ADMs, and each ADM is defined by the following class prototype:

\begin{footnotesize}\begin{verbatim}
template<class DataStructure, class Selector, 
         class Migration, class NextADM>
class AtomicDMM {
  ...
  inline void* malloc (size_t sz) { ... }
  inline void  free   (void* ptr) { ... }
  ...
};
\end{verbatim}\end{footnotesize}
where
\begin{itemize}
\item \emph{DataStructure} is the data structure of the ADM designed for a certain region of memory. It should include the type of data structure and policies for blocks sorting and selection that are used in that manager (trees A.1, A.3, A.4 and B.1 in Fig. \ref{fig:OrthogonalTrees}).
\item \emph{Selector} includes the set of conditions determining the range of block sizes that will be attended by this ADM. If there are several ADMs with the same range, every memory request is attended in descending order as the ADMs are created in the code, in such a way that the last ADM attends requests when there are no free blocks on the previous atomic DMMs (tree A.2).
\item \emph{Migration} defines the set of rules determining the range of block sizes that are returned (freed) by this atomic DMM. Using this parameter, block migration policies between different ADMs can be defined, that is, coalescing and splitting policies (trees A.5, C.1, C.2, D.1 and D.2 in Fig. \ref{fig:OrthogonalTrees}).
\item \emph{NextADM} is the next ADM in the global manager' structure. If there are no more ADMs, it represents the interface used by the \emph{Operating System (OS)} to de/allocate memory (\emph{sbrk()}, \emph{mmap()}, \emph{malloc()}, etc.).
\end{itemize}

For illustration purposes, in the following example we design an excerpt of the Kingsley DMM (see Table \ref{tab:Kingsley}). As stated before, this DMM is formed by up to 30 ADMs. Thus, such DMM manages 30 different regions of memory. Every region is selected according to the block size that the application needs to de/allocate. The first ADM uses a singly-linked list of blocks with \emph{First Fit (FF)} allocation policy (\emph{FirstFitSLL} data structure). This atomic manager attends de/allocation for 2$^3$-bytes-size objects, and does not use coalescing or splitting (\emph{SizeSelector} may be used as both de/allocation size and migration policy). The second atomic manager implements the same behavior, although it is used for 2$^4$-bytes-size objects, and so forth. Finally, the last region is used for all the requests that cannot be managed by the previous 30 atomic managers.

\begin{footnotesize}\begin{verbatim}
typedef KingsleyDMM <
AtomicDMM<
 FirstFitSLL<SizeHeader>,
 SizeSelector<8>,
 SizeSelector<8>,
 AtomicDMM<
  FirstFitSLL<SizeHeader>,
  SizeSelector<16>,
  SizeSelector<16>,
  ...
  AtomicDMM<
   FirstFitSLL<SizeHeader>,
   SizeSelector<2^32>,
   SizeSelector<2^32>,
   OperatingSystem<SbrkHeap<EmptyHeader>, 
             2048KB, SizeHeader>
  >
 >
>> GlobalHeap;
\end{verbatim}\end{footnotesize}

Finally, we have also extended this DMM library with a {\it simulation mode}. It allows us not only to execute a real DMM for a certain application, but also to emulate the behavior of a DMM to obtain the performance, the memory used and energy consumed by the embedded application in independent cases or memory allocation situations. In simulation mode, the library does not de/allocate memory from the computer like the real application, but maintains useful information about how the structure of the selected DMM evolves in time. Thus, we are able to evaluate a DMM with an initial profiling of the application faster than previous approaches, where every DMM is evaluated running the application in real time with a predefined DMM. As a result, the evaluation of DMM can be performed relatively fast, and exploration algorithms can be included in the searching process.

\subsection{Grammatical evolution applied to DMM optimization}

\emph{Grammatical Evolution (GE)}~\cite{ONeill2003,Brabazon2006,Dempsey2007,ONeill2001,Brabazon2008} is a grammar-based form of \emph{Genetic Programming (GP)} \cite{Poli2008}. It combines principles from molecular biology to the representation power of formal grammars. GE's rich modularity gives a unique flexibility, making it possible to use alternative search strategies (evolutionary, deterministic or some other approach) and to radically change its behavior by merely changing the supplied grammar. Since a grammar describes the structures that are generated by GE, it is trivial to modify the output structures by simply editing the plain text grammar. This is one of the main advantages that makes the GE approach so attractive. The genotype-phenotype mapping also means that instead of operating exclusively on solution trees, as in standard GP, GE allows search operators to be performed on the genotype (e.g., integer or binary chromosomes), in addition to partially derived phenotypes and the fully formed phenotypic derivation trees themselves. When tackling a problem with GE, a suitable \emph{Backus Naur Form (BNF)} grammar definition must be initially defined. The BNF can be either the specification of an entire language or, perhaps more usefully, a subset of a language geared towards the problem at hand. In a simulation run, GE can theoretically evolve programs in any language described by a BNF.

A grammar can be represented by the tuple $\left\{N, T, P, S\right\}$, where $N$ is the set of non-terminals, $T$ the set of terminals, $P$ a set of production rules that maps the elements of $N$ to $T$, and $S$ is a start symbol that is a member of $N$. When there are a number of productions that can be applied to one element of $N$, the choice is delimited with the ``$|$'' symbol.

\begin{figure}[!t]
\centering
\includegraphics[width=0.60\columnwidth]{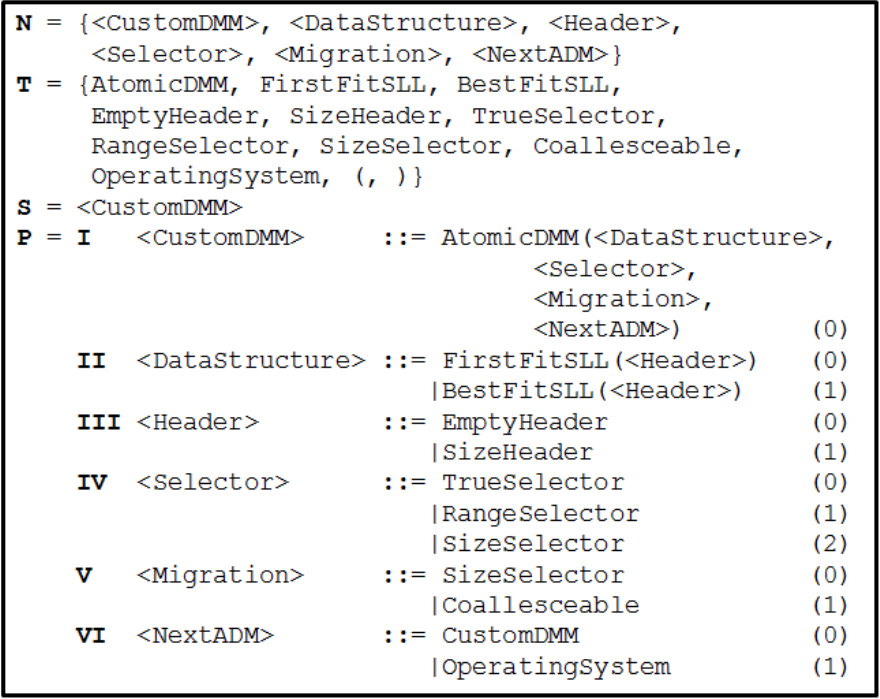}
\caption{Illustrative grammar describing a set of DMM implementations.}
\label{fig:GrammarExample}
\end{figure}

\begin{figure}[!t]
\centering
\includegraphics[width=0.60\columnwidth]{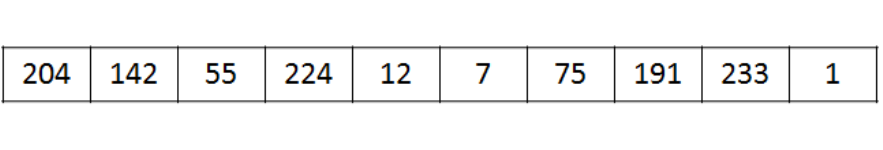}
\caption{A GE individual's genome.}
\label{fig:GenotypeExample}
\end{figure}

A GE's individual uses a variable-length encoding scheme where each gene holds an integer value that will be mapped to previously labeled production rules of a given BNF by the decoding process. The genotype is used to map the start symbol, as defined in the grammar, onto terminals by reading codons to generate a corresponding integer value. For illustration purposes, in Fig. \ref{fig:GrammarExample} we shown an excerpt of a simple grammar that defines a language for DMM implementations, including two data structures (\texttt{FirstFitSLL} and \texttt{BestFitSLL}), two headers, three selectors, two policies to manage flexible or fixed block sizes, and the main memory managed by the OS (\texttt{OperatingSystem}). Fig. \ref{fig:GenotypeExample} illustrates a genome  which has 10 genes with values ranging from 0 to 255 (8-bit number). Since an 8-bit integer is far more than the number of production rules, the modulus operation is needed to decode the genes properly.

The decoding process matches each gene with the corresponding production rules group. The first gene, namely 204, corresponds to the group denoted as \textbf{I}. Since there is only one production rule headed by \texttt{<CustomDMM>}, the selected one is the production labeled 0 (204 mod 1 = 0) (\texttt{<CustomDMM> ::= AtomicDMM(<DataStructure>, <Selector>, <Migration>, <NextADM>)}). Next, the second gene is read and its value (142), after the modulus operation (142 mod 2), results in 0; therefore, the production \texttt{<DataStructure> ::= FirstFitSLL(<Header>)} is picked up from the group denoted as \textbf{II}. Next, there are two productions pointed to by the symbol \texttt{<Header>}, so the gene 55, after the modulus operation (55 mod 2) results in 1 and the rule \texttt{<Header> ::= SizeHeader} is selected. Thus, the decoded expression is:

\begin{footnotesize}\begin{verbatim}
AtomicDMM(FirstFitSLL(SizeHeader), 
          <Selector>, 
          <Migration>, 
          <NextADM>)
\end{verbatim}\end{footnotesize}

The next two steps will sequentially produce the following expressions:

\begin{footnotesize}\begin{verbatim}
AtomicDMM(FirstFitSLL(SizeHeader), 
          SizeSelector, 
          <Migration>, 
          <NextADM>)
AtomicDMM(FirstFitSLL(SizeHeader), 
          SizeSelector, 
          SizeSelector, 
          <NextADM>)
\end{verbatim}\end{footnotesize}
and finally, at the 6th gene, the full decoded expression will be:

\begin{footnotesize}\begin{verbatim}
AtomicDMM(FirstFitSLL(SizeHeader), 
          SizeSelector, 
          SizeSelector, 
          OperatingSystem)
\end{verbatim}\end{footnotesize}

An important advantage of this grammatical evolution representation is that it uses a linear genome. Therefore, GE can directly use all standard genetic algorithm operators. Furthermore, because of the simplicity of the linear representation, computing implementations of GE are relatively simple to deploy. In particular, as shown in the previous example, the DMM represented by an individual can be deployed in $O(N)$, where $N$ is the number of genes.

In the previous example the decoding process terminated without translating all genes. This occurs because  the genetic operations in GE do not know about the semantics of a genome until it is decoded. Then, the decoding process frequently ends up with a complete expression (final) without traversing the entire genome. Hence, this exploration process has a short execution time.

However, a potentially serious concern with GE is the situation of having redundant genes. During the decoding process, it may happen that a genome does not have enough genes to generate a complete expression (sentence), i.e., there are still non-terminals remaining. In this case the \emph{wrap} operator is applied, which results in returning the codon reading head back to the first codon in the individual. Hence, even after wrapping, the mapping process would be incomplete and would carry on indefinitely unless terminated. This occurs because a non-terminal is being mapped recursively by a production rule. Such an individual is labeled as invalid, since it will never undergo a complete mapping to a set of terminals. For this reason our optimization approach for DMM imposes an upper limit on the number of wrapping events that can occur. Therefore, during the mapping process, starting from the left-hand side of the genome, integer values are generated and used to select rules from the BNF grammar, until one of the following situations arises:

\begin{enumerate}
	\item A complete DMM is generated. This occurs when all the non-terminals in the expression being mapped are transformed into elements from the terminal set of the BNF grammar.
	\item The end of the genome is reached. In this case the wrapping operator is invoked returning the genome reading frame to the left hand side of the genome. The reading of codons will then continue, unless the upper threshold value representing the maximum number of wrapping events, during the individual's mapping process, is reached.
	\item In case the threshold value on the number of wrapping events is reached and the individual is still incompletely mapped, then the mapping process is halted, and the highest possible fitness\footnote{In evolutionary algorithms, the fitness is the measure of quality for a given individual.} value (assuming minimization) is assigned to the individual.
\end{enumerate}

In summary, to evolve different DMMs in an optimization process, we have defined a grammar which covers any possible DMM that our template library can instantiate. Moreover, our grammar is complete enough to implement any well-known DMM and to explore custom DMM implementations for real-life multimedia embedded applications. Moreover, although this methodology was originally developed to optimize embedded system designs (which introduce strong performance, memory and energy constraints), it can be applied to any computer platform where the use of C++ is allowed.

\begin{figure}[!t]
\centering
\includegraphics[width=0.45\columnwidth]{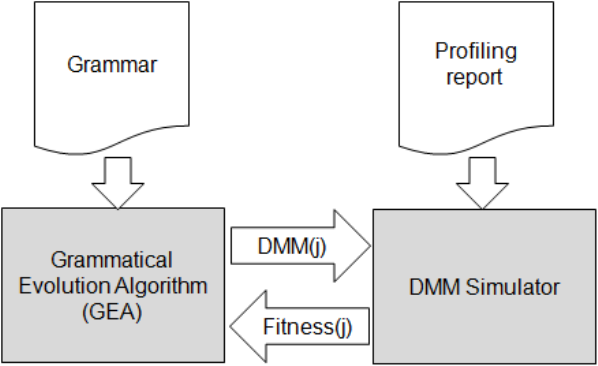}
\caption{DMM generation and evaluation process.}
\label{fig:Loop}
\end{figure}

Finally, Fig. \ref{fig:Loop} shows an illustrative example on how our methodology performs. The main inputs are a profiling report, which logs all the blocks that have been de/allocated, and the grammar file. Our GE algorithm is constantly generating different DMM implementations from the grammar file. When a DMM is generated ($\mathit{DMM(j)}$ in Fig. \ref{fig:Loop}), it is received by the DMM library working the \emph{simulation mode} (referred as DMM simulator in Fig. \ref{fig:Loop}) . Next, the DMM simulator emulates the behavior of the application, debugging every line in the profiling report. Such emulation does not de/allocate memory from the computer like the real application, but maintains useful information about how the structure of the selected DMM evolves in time. After the profiling report has been simulated, the DMM library returns back the fitness of the current DMM to the GE algorithm.


\section{DMM optimization flow}
\label{sec:OptimizationFlow}

\begin{figure}[!t]
\centering
\includegraphics[width=0.75\columnwidth]{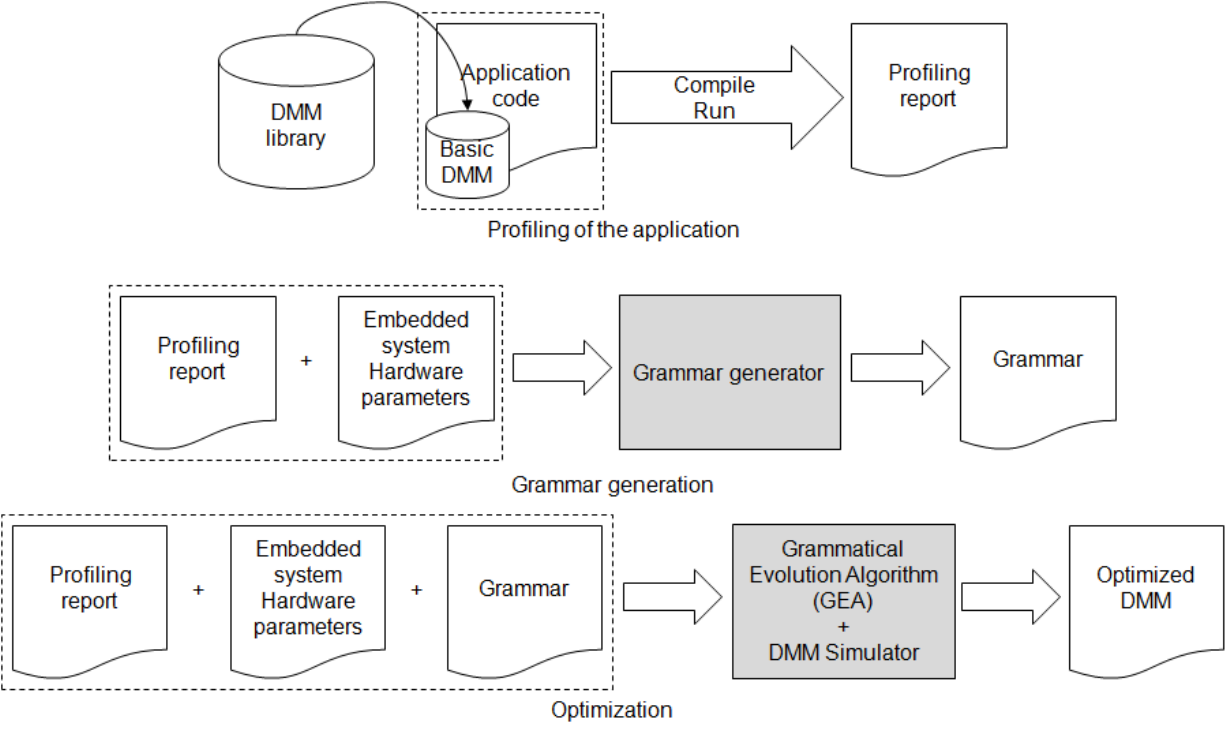}
\caption{DMMs optimization flow. In the first phase, we generate an initial profiling of the de/allocation pattern. In the second phase, we automatically analyze the profiling report and provide the hardware parameters of the target embedded system to generate the final grammar. Finally, in the third phase an exploration of the design space of DMMs implementation is performed using GE.}
\label{fig:Framework}
\end{figure}

The proposed optimization framework uses three different phases to perform the automatic exploration of DMMs using grammatical evolution. Fig. \ref{fig:Framework} shows the different phases required to perform the overall DMMs optimization. In the first phase, we generate an initial profiling of the de/allocation pattern of the different objects instantiated by the application. In the second phase, we automatically analyze the profiling report and provide the hardware parameters of the target embedded system to generate the final grammar, customized for the application and final embedded system under study. Finally, in the third phase an exploration of the design space of DMMs implementation is performed using GE. Next, we detail the three phases of our proposed optimization flow.

\subsection{Profiling of the application}

First, we run the application under study using a basic DMM implemented with the DMM library. Such process logs all the required information in an external file: identification of the object created/deleted, operation (allocation or deallocation), object size in bytes and memory address. To this end, we must only include the DMM library in the source code of the application (only one line of code at the beginning of the application to optimize). As a result, in our methodology, this phase takes between 30-45 minutes for real-life applications, with very limited user interaction. Since all the exploration process is performed simulating the generated DMMs with the profiling report, this task is done just once.

\subsection{DMM grammar generation}

According to Fig. \ref{fig:Framework}, all the information contained in the profiling report is automatically examined and a customized grammar for the memory architecture of the target system is obtained.
As a result, some incomplete rules in the original grammar, such as the memory size of the embedded system, are automatically defined according to each specific profiling. To this aim, a tool called \emph{Grammar Generator} has been developed. This phase takes no more than a couple of minutes, and without user interaction.

\subsection{Optimization}
\label{sec:optimization}
The last phase is the optimization process. As Fig. \ref{fig:Framework} depicts, this phase consists of a GE algorithm that takes as input: (1) the grammar generated in the previous phase, (2) the hardware parameters (e.g., memory size and power consumption model for the embedded memory~\cite{Mamidipaka2004}) of the target embedded system, and (3) the profiling report of the application. This phase also uses the simulation mode extension of the DMM library in order to obtain the fitness of every DMM generated by the grammar.

The fitness is computed as a weighted sum of the performance, memory usage and energy consumed by the proposed DMM for the target embedded system and application under study. Such parameters are indirectly calculated by the DMM simulator as follows: it calculates the computational complexity or time complexity \cite{Sipser2005} to compute performance, the size of memory de/allocated by the DMM to compute memory usage, and the number of memory accesses to compute energy. In this regard, every portion of the code in the simulator that emulates the behavior of a DMM is accompanied by its corresponding added execution time, memory accesses and memory usage. The following code snippet shows an illustrative example of how this task is performed:

\begin{footnotesize}\begin{verbatim}
inline void malloc(size_t sz) {
 exTime += 2; memAcc += 2;
 object* ptr = head.next;
 if(ptr!=&tail) {
  exTime += 2; memAcc += 5;
  head.next = ptr->next;
  if(head.next==&tail) {
   exTime++; memAcc += 2;
   memUsed -= ptr->size();
   tail.next = &head;
  }
  exTime++;
  return (void*)ptr;
 }
 exTime++;
 return 0;
}
\end{verbatim}\end{footnotesize}
The first sentence computes the execution time of the pointer assignment (\texttt{ptr}) and the evaluation of the \texttt{if} condition. The second one takes into account two memory accesses: one for the \texttt{head.next} sentence (i.e., access operator) and one because of the \texttt{\&tail} sentence. This process is repeated until the end of the function, updating the execution time, memory accesses and memory used when needed. In the example presented, the memory used is reduced in \texttt{ptr->size()}: this is correct because the DMM does not need to manage this portion of memory, unless it is freed. Finally, the energy consumed is computed using the total number of memory accesses.

When the optimization process ends, the GE algorithm returns the best DMM found, with minimal weighted sum of execution time (performance), memory usage and energy consumption. In the two benchmarks tested, this phase varies from 10 to 16 hours with no user interaction mainly depending on the size of the profiling report. In our experiments we have applied GE to profiling reports varying from 3 to 5 GB. Note that in previous approaches, this phase typically takes days or weeks, and for every DMM generated the application must be compiled and executed to evaluate the fitness function. In any case, our methodology requires much less time than state-of-the-art solutions to this problem because it processes a profiling report, instead of simulating multiple times the complete original application. Furthermore, we do not compile the original application every time a new DMM must be evaluated, which makes our framework even more stable and overall results more easily comparable. As a consequence, our (sequential) methodology is up to 24$\times$ faster than previous approaches proposed in the literature (\cite{Berger2001}, \cite{Atienza2006a}), and allows the system designer to use automatic exploration algorithms.

In the optimization of DMMs, we address two fundamental problems related to the degree of minimization (quality of solutions) and the exploration time. Regarding the quality of solutions \cite{deb2001}, since it is not possible to obtain the true optimal due to the exponential exploration time with respect to the possible DMM solutions, we use GE to obtain good approximations. Moreover, in the following section we show that a suitable parallel implementation can improve the quality of solutions. With respect to exploration time, current multimedia applications include thousands of dynamic memory operations and obtaining the optimal solution is a very time-consuming problem, even if we improved the evaluation of candidate DMMs by means of simulation. Hence, our proposed design of a \textit{parallel GE Algorithm (pGEA)} for DMMs optimization can effectively reduce the exploration time when more processor cores are used. 


\section{pGE parallelization approach for scalable DMM exploration}
\label{sec:ParallelImplementation}

\subsection{Global architecture of the pGE}

The proposed search process can be significantly improved by using several threads to perform the simulations instead of running each one in the same processor. In this section we first describe the different approaches used in the parallelization of evolutionary algorithms, selecting one of them. Finally we analyze how our GE algorithm works in a parallel environment to solve the exploration of DMMs in embedded applications. 

There are two main levels at which an evolutionary algorithm can be parallelized: the fitness evaluation level and the population level \cite{Fernandez2003}. In the following we briefly describe both models. For a more detailed discussion see \cite{Fernandez2005}.

\begin{figure}[!t]
\centering
\includegraphics[width=0.45\columnwidth]{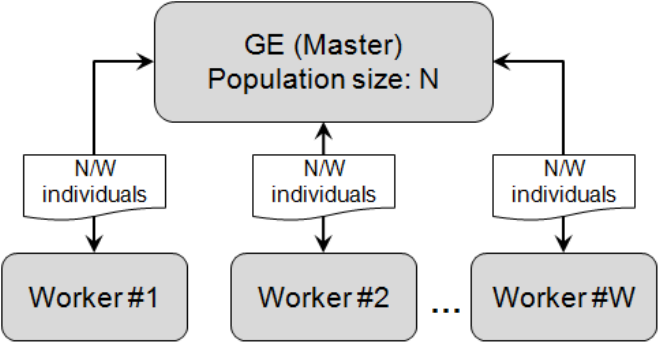}
\caption{Master-worker scheme.}
\label{fig:MasterWorker}
\end{figure}

In the fitness evaluation level, the calculation of the individuals' fitness is by far the most time consuming step of the evolutionary algorithm. In such cases an obvious approach is to share fitness evaluation among several processors. One example is the master-worker algorithm: this evolutionary algorithm maintains one population, while the evaluation of fitness is distributed among several processors (Fig. \ref{fig:MasterWorker}). As in the serial algorithm, selection and mating are global: each individual may compete and mate with any other. The evaluation of the individuals is usually parallelized because the fitness of an individual is independent from the rest of the population and there is no need to communicate during this phase. Communication occurs only as each worker receives its subset of individuals to evaluate and when the workers return the fitness values. This parallel algorithm is synchronous since it stops and waits the fitness values for all the population before proceeding into the next generation. As a result, it has the same properties as the sequential ones, but it is faster if the algorithm spends most of the time for the evaluation process.

Regarding the population level, natural populations tend to possess a spatial structure and to be organized in so-called demes. Demes are semi-independent groups of individuals or subpopulations which have only a loose coupling to other neighboring demes. This coupling takes the form of the migration of individuals between demes. Several models based on this idea have been proposed. The two most important are the island and the grid models. In the island model individuals are allowed to migrate between populations with a given frequency. Several patterns of exchange have been traditionally used. The most common ones are rings, two-dimensional and three-dimensional meshes, stars and hypercubes. The most common replacement policy, consists of replacing the worst $k$ individuals in the receiving
population with $k$ immigrants which are the best $k$ individuals of their original
island \cite{Hidalgo2007}. In the grid or cellular model, individuals are placed in a
one or two-dimensional grid, one individual per grid location. The genetic operations take place locally, in a small neighborhood of a given individual. Although the model can be implemented on a standard computer, an efficient distributed implementation is obtained by partitioning the whole population into subpopulations, where each subpopulation is a grid of individuals. This model has a straightforward implementation on a cluster.

Given these three different algorithms, we propose the use of a synchronous master-worker \emph{parallel GE Algorithm (pGEA)} where the master applies GE operators and each worker runs a different set of DMM simulations (or fitness evaluations). Synchronous master-worker schemes have many advantages: (1) they explore the search space exactly as a sequential algorithm and thus we are able to compare the parallel algorithm with our previous sequential algorithm presented in \cite{RiscoMartin2009b}, (2) they are easy to implement and (3) significant performance improvements are possible in many cases \cite{CantuPaz1998b}. Moreover, the master-worker topology can be easily implemented using the DEVS formalism \cite{Zeigler2000}.

Fig. \ref{fig:MasterWorker} shows the parallel process. There is one master node and $W$ workers. The master node executes all the GE operations except the fitness evaluation. In the evaluation step, a set of $\frac{N}{W}$ individuals are sent to every worker, where $N$ is the total population size. The work load is balanced among all the workers by applying a round-robin algorithm: we distribute the individuals according to an estimation of the simulation time, which is computed based on the number of ADMs that the DMM contains. This metric is not accurate at the first generations of the algorithm. However, as the optimization algorithm progresses, the allocation policies are homogeneous and the simulation time highly depends on this metric. Finally, the master waits for the results computed by the workers. So, in the best case, if the algorithm spends most of the time for the evaluation process, the total execution time is reduced in a $\frac{T}{W}$ factor, where $T$ is the execution time of the serial algorithm. Note that in our optimization of embedded systems, the evaluation of the different DMMs is about 98\% of the entire computational time.

The algorithm in Fig. \ref{fig:MasterWorker} follows a distributed or multi-threaded design, which is suitable to be executed in multi-core architectures, as well as workstations connected over a LAN. The approach we have implemented consists of executing our proposed pGEA in a set of PCs connected over a LAN.

To this aim, we propose to implement our pGEA in a distributed DMMs exploration using \emph{Discrete Event System Specification (DEVS)}, which is a theory of modeling and simulation recently proposed. In the following, we provide a formal description of DEVS, our DEVS/SOA model and the algorithm needed to implement our pGEA. 

\subsection{DEVS/SOA parallel exploration algorithm}

DEVS is a general formalism for discrete event system modeling based on set theory \cite{Zeigler2000}. Once a system is described in terms of the DEVS theory, it can be easily implemented using an existing library. To better understand this point, we may show a couple of analogies: (1) in control theory, when a system is described using block diagrams, it can be easily implemented using Matlab, Scilab or Dymola, and (2) in the field of software engineering, a system may be implemented using object-oriented programming when it is documented in terms of class diagrams and state machines. 

There are now numerous libraries and tools for expressing DEVS models across the globe, such as DEVSJAVA, xDEVS, DEVS/C++, CD++, aDEVS, DEVS-Suite, James, etc \cite{Mittal2009a}. Although this proliferation of libraries shows the numerous advantages in doing DEVS M\&S, its multiplicity presents a difficulty in sharing models. In this regard, we have developed a new DEVS simulation framework based on web services called \emph{Discrete Event Systems Specification over Service Oriented Architecture (DEVS/SOA)} \cite{Mittal2009a}, which we have made available at~\cite{RiscoWeb2009}. The main advantage of using DEVS/SOA is that the original (sequential) model may be distributed with no additional parallelization middleware support, i.e., the whole system can be distributed using a standard DEVS library. Another major advantage is that DEVS/SOA allows the engineer to combine several DEVS platforms to model a system, i.e., it provides interoperability between multiple (and distributed) processing architectures. It does not happen with other parallelization frameworks like MPI \cite{Gropp1999}, \cite{Lima2009} or HLA \cite{HLA2000}, where the software engineer must develop an extra layer in the sequential algorithm in order to provide such level of parallelization.

DEVS/SOA enables the representation of a distributed system by three sets and four functions: input set $(X)$, output set $(Y)$, state set $(S)$, external transition function ($\delta_{ext}$), internal transition function ($\delta_{int}$), confluent function ($\delta_{con}$), and output function ($\lambda$). The DEVS formalism (and by extension our proposed DEVS/SOA formalism) provides the framework for information modeling which gives several advantages to analyze and design complex systems: completeness, verifiability, extensibility, and maintainability. Unlike DEVS, which presents two kind of models i.e. atomic and coupled \cite{Zeigler2000}, DEVS/SOA has one kind of model to represent systems: the atomic model, that can specify complex systems in a hierarchical way. DEVS/SOA model processes an input event based on its state and condition, and it generates an output event and changes its state. Finally, it sets the time during which the model can stay in that state.

Formally, a DEVS/SOA model is a set of atomic models, and each atomic model is defined by the following equation:

\begin{equation}
A=\langle IP, OP, X, S, Y, \left\{ A_{i}\right\} , C, \lambda, \delta_{ext}, \delta_{int}, \delta_{con}\rangle
\label{eq:DevsModel}
\end{equation}
Where:
\begin{itemize}
\item $IP$ is the set of input ports.
\item $OP$ is the set of output ports.
\item $X$ is the set of inputs described in terms of pairs port-value: $\left\{ p,v\right\} $.
\item $S$ is the state space. It includes not only the current state of the atomic model, but also two special parameters called $\sigma$ and \emph{phase}, which is the time until the next event generation, and a description of the current state (usually in natural language), respectively.
\item $Y$ is the set of outputs, also described in terms of pairs port-value: $\left\{ p,v\right\} $.
\item $A_{i}$ is the set of atomic models that the current model $A$ contains. Note that $A_{i}$ makes this definition recursive.
\item $C$ is the set of connections from the current atomic model $A$ to others.
\item $\lambda$ is the output function. When the time passed since the last output function is equal to $\sigma$, then $\lambda$ is automatically executed.
\item $\delta_{int}$ is the internal transition function. It is executed right after the output ($\lambda$) function and is used to change the state $S$ (including \emph{phase} and $\sigma$)
\item $\delta_{ext}$ is the external transition function. It is automatically executed when an external event arrives to one of the input ports, changing the current state if needed.
\item $\delta_{con}$ is the confluent function, and is selected if $\delta_{ext}$ and $\delta_{int}$ must be executed at the same time instant.
\end{itemize}

In this regard, we may describe both master and worker as DEVS atomic models. 

\subsubsection{Master model}

\begin{figure}[t]
\centering
\includegraphics[width=0.50\columnwidth]{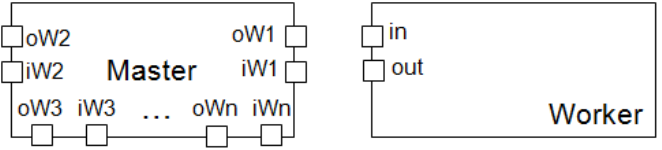}
\caption{Master (left) and Worker (right) DEVS model structure.}
\label{fig:MasterWorkerDevs}
\end{figure}

The master runs the GE algorithm, and includes a set of pairs $\left\{ {iW}_j, {oW}_j \right\}, j \in 1, 2, 3 \ldots W$, for input and output ports. \emph{Out} connections are used to send the subset of $\frac{N}{W}$ DMMs to the corresponding worker, to be simulated and evaluated. \emph{In} connections are used to receive the results of such simulations. Fig. \ref{fig:MasterWorkerDevs} depicts the structure of the master atomic model. Algorithm \ref{alg:DevsSoaMasterImpl} shows our DEVS/SOA atomic algorithm that implements the master node. As stated above, this DEVS/SOA model must contain the GEA, sending the subset of DMM implementations to the corresponding neighbor.

\begin{algorithm}[!t]
\caption{Master node: DEVS/SOA atomic implementation}
\begin{algorithmic}

\REQUIRE{GEA is the implementation of our evolutionary algorithm. $W$ is the number of workers. $N$ is the population size. \emph{received} is a state variable and is \emph{true} if the current atomic model has received the DMM evaluations by all the workers. }

\vspace{0.25cm}
\textbf{function} init()
\STATE init(GEA) \COMMENT{Initialization of the GEA, the grammar is loaded and the first random population of DMMs is generated}
\STATE $t=0$
\STATE \emph{phase = ``active''}
\STATE $\sigma = 0$

\vspace{0.25cm}
\textbf{function} $\lambda ()$
\IF{\emph{phase == ``active''}}
 \STATE sort(GEA) \COMMENT{sort individuals to balance the load among all the workers. An estimation of the simulation time is performed.}
 \FOR{$j=1$ to $W$}
  \STATE send(``oW$_j$'', subpop(GEA, $j$, $W$)) \COMMENT{sends $\frac{N}{W}$ DMMs to worker $j$ by the ``oW$_j$'' output port.}
 \ENDFOR
\ENDIF

\vspace{0.25cm}
\textbf{function} $\delta_{int}()$
\STATE \emph{received = false}
\STATE \emph{phase = ``passive''}
\STATE $\sigma = \infty$

\vspace{0.25cm}
\textbf{function} $\delta_{ext}()$
\FOR{$j=1$ to $W$}
 \IF{not empty(``iW$_j$'')}
  \STATE receive(``iW$_j$'', GE, $j$, $W$, \emph{received}) \COMMENT{Updates the current population with the evaluated DMMs coming from port ``iW$_j$''. It also updates the \emph{received} state variable, which will be true when all the workers have evaluated their DMMs}
 \ENDIF
\ENDFOR
\IF{$t < T$ and \emph{received == true}}
 \STATE step(GEA) \COMMENT{Computes the offspring of the current generation}
 \STATE $t = t +1$
 \STATE \emph{phase = ``active''}
 \STATE $\sigma = 0$
\ENDIF

\vspace{0.25cm}
\textbf{function} $\delta_{con}()$
\STATE $\delta_{int}()$
\STATE $\delta_{ext}()$

\end{algorithmic}
\label{alg:DevsSoaMasterImpl}
\end{algorithm}

Taking into account Algorithm \ref{alg:DevsSoaMasterImpl}, the proposed master model works as follows. First, at the \emph{init} function, GEA is initialized and both phase and $\sigma$ are set to ``active'' and $0$, respectively. It means that the master immediately executes the output function $\left(\lambda\right)$, sending the DMMs to the workers. After the output function, the internal transition function is executed $\left(\delta_{int}\right)$, which passivates the master waiting for the results $\left(\sigma=\infty\right)$ . These results are gradually sent by the workers and received in the input ports. When a subset of DMMs arrives, the external transition function $\delta_{ext}$ is automatically executed. It first updates the evaluated DMMs of the current population and then updates the \emph{received} variable, which is \emph{true} when all the DMMs have been evaluated at the workers and received by the master. When this condition is met, the master model is ready to compute the offspring and restart the process. To this end, $\sigma$ is set to $0$ and the output function is executed again. This procedure is repeated until the maximum number of generations is reached ($t < T$ condition in the external transition), although other conditions can be used, for example when a certain value for the fitness is reached. After that, $\sigma$ is set to infinity and the simulation ends.

\subsubsection{Worker model}

Every worker has one pair of $\left\{ in, out \right\} $ ports (Fig. \ref{fig:MasterWorkerDevs}), to receive and to send the subset of DMMs sent and received by the master, respectively.

Similarly, the worker atomic model is presented in Algorithm \ref{alg:DevsSoaWorkerImpl}. It works as follows. First, in the \emph{init} function, the DMM simulator is initialized, loading the profiling report (every worker keeps a copy of this file). Both phase and $\sigma$ are set to ``passive'' and $\infty$, respectively, because the worker starts to evaluate DMMs just when set of DMMs is received by the {\tt in} input port. When this happens, the external transition function is executed $\left(\delta_{ext}\right)$. It means that the master has sent the set of DMMs to be evaluated. Thus, the simulation starts, and every DMM received is simulated, computing its fitness. After that, both phase and $\sigma$ are set to ``active'' and $0$ respectively. Then, the output function $\left(\lambda\right)$ is automatically executed and the updated DMMs are sent back to the master through the {\tt out} output port. After the output function, the internal transition function is executed $\left(\delta_{int}\right)$, passivating the model until a new set of DMMs is received.

\begin{algorithm}[!t]
\caption{Worker node: DEVS/SOA atomic implementation}
\begin{algorithmic}

\REQUIRE{DMMSimulator is the simulator. \emph{dmms} is the set of DMMs sent by the master node.}

\vspace{0.25cm}
\textbf{function} init()
\STATE init(DMMSimulator) \COMMENT{Initialization of the simulator, the profiling report is loaded}
\STATE \emph{phase = ``passive''}
\STATE $\sigma = \infty$

\vspace{0.25cm}
\textbf{function} $\lambda ()$
\IF{\emph{phase == ``active''}}
 \STATE send(``out'', \emph{dmms}) \COMMENT{sends the evaluated DMMs to the master by the ``out'' output port}
\ENDIF

\vspace{0.25cm}
\textbf{function} $\delta_{int}()$
\STATE \emph{phase = ``passive''}
\STATE $\sigma = \infty$

\vspace{0.25cm}
\textbf{function} $\delta_{ext}()$
\IF{not empty(``in'')}
 \STATE receive(``in'', \emph{dmms}) \COMMENT{The set of DMMs sent by the master is loaded into the \emph{dmms} state variable}
 \STATE evaluate(DMMSimulator, \emph{dmms}) \COMMENT{The DMMSimulator computes the fitness of each DMM received}
 \STATE \emph{phase = ``active''}
 \STATE $\sigma = 0$
\ENDIF

\vspace{0.25cm}
\textbf{function} $\delta_{con}()$
\STATE $\delta_{int}()$
\STATE $\delta_{ext}()$

\end{algorithmic}
\label{alg:DevsSoaWorkerImpl}
\end{algorithm}

\subsubsection{DEVS/SOA configuration}

Our DEVS/SOA implementation establishes that each atomic model can be placed in the same processor (single-core), in different processors of the same computer (multi-core) or in different processors at different computers (distributed). The main advantage of using DEVS/SOA is that the original model remains unchanged under all possible configurations.

\begin{figure}[t]
\centering
\includegraphics[width=0.80\columnwidth]{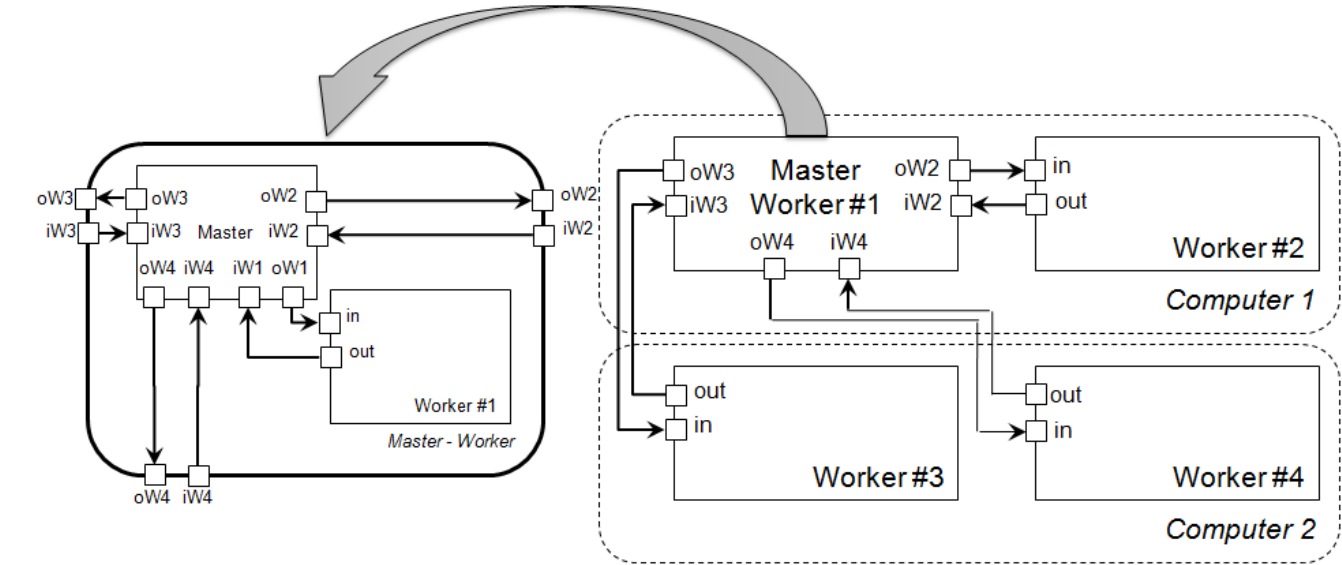}
\caption{Graphic representation of our pGEA topology using the DEVS/SOA formalism. The figure on the left depicts the complex master-worker atomic model, formed by one master and one worker atomic models. The figure on the right represents the top level model, one master and four workers.}
\label{fig:DEVSpGEA}
\end{figure}

Since we are exploring DMMs in 2-core computers, it is desirable to place one atomic model per processor to reach good performance. Thus, one possible DEVS/SOA topology is formed by two atomic models running one worker on each one of them. At the top, the master model can be located with a single worker. However, as it has been stated above, the master node processes only the 2\% of the total load, so we may conclude that placing a single master in a single core is not profitable. As a conclusion, we may place one worker per core, but one of these cores must contain both a master and a worker as a complex atomic model. Such complex atomic model is named \emph{master-worker}. Fig.~\ref{fig:DEVSpGEA} provides a scheme of our pGEA design in two computers. Each atomic model contains either one master-worker or worker, and there must be just one master-worker in the whole distributed model. This configuration is performed in DEVS/SOA just writing a single \textit{XML (eXtensible Markup Language)} file. Further details on how to deploy such file can be found in \cite{Mittal2009a}.

The DEVS/SOA approach we have implemented consists of executing our proposed pGEA solution for DMMs exploration in a set of workstations connected over a local area network. To this end, using our DEVS/SOA framework, we can distribute atomic models over a variable number of workstations. The pGEA's configuration enables scaling both performance and quality of solutions when different number of processors are available to execute the exploration framework. 


\section{Experimental framework and results}
\label{sec:Experiments}

We have evaluated the proposed optimization framework for two multimedia embedded applications~\cite{VDrift,Kharevych2002}. Thus, in this section we describe the complete method applied to compare the different type of sequential and parallel GEAs while optimizing dynamic embedded application, as well as the experimental results obtained.

\subsection{Case studies}
\label{subsec:CaseStudies}

The first case study is VDrift~\cite{VDrift}, which is a driving simulation game. The game includes as main features: 19 different tracks, 28 types of cars, artificial intelligent players and a networked multi-player mode. The second benchmark is a 3D Physics Engine (Physics3D) for elastic and deformable bodies \cite{Kharevych2002}, which is a 3D engine that displays the interaction of non-rigid bodies.

To implement our GEA, we have used GEVA \cite{ONeill2008}, a well-known GE tool developed in Java. The distributed pGEA version is made by adding the DEVS/SOA framework, which will use multiple processors when available. The pGEA configuration based on DEVS/SOA was distributed on a set of 2 nodes Intel\textsuperscript{\textregistered} Core{\texttrademark} 2 CPU 6600 2.40GHz with 2GB DDR memory, connected via a 100Mbps Ethernet network. To this end, we placed two DEVS/SOA atomic models (following Fig. \ref{fig:DEVSpGEA}) per node. This leads to 4 atomic DEVS/SOA models running in parallel. All the values presented are computed by averaging results of 10 trials, according to a MonteCarlo simulation.

\subsection{DMMs exploration speed comparisons}

First, we compare the execution time employed by our pGEA to explore optimal DMMs with two sequential methodologies. The first one was proposed by Atienza \emph{et al} \cite{Atienza2006} and has been described in Section \ref{sec:Introduction}. The second one is our original GEA. Table \ref{tab:Parameters} shows the parameters used by both GEA and pGEA. Moreover, we have tested four different parallel configurations with 1, 2, 3, and 4 workers (called pGEA$_1$, pGEA$_2$, pGEA$_3$ and pGEA$_4$, respectively). To configure the different parallel environments, we just set the $W$ parameter of the master-worker model to 1, 2, 3 and 4, respectively.

\begin{table}[ht]
\caption{Parameters for both GEA and pGEA.}
\begin{center}
\begin{tabular}{|l|l|}
\hline
\textbf{Parameter} & \textbf{Value} \\
\hline
Population size & 60 \\
\hline
Number of generations & 100 \\
\hline
Probability of crossover & 0.80 \\
\hline
Probability of mutation & 0.02 \\
\hline
\end{tabular}
\end{center}
\label{tab:Parameters}
\end{table}

\begin{figure}[!t]
\centering
\includegraphics[width=0.50\columnwidth]{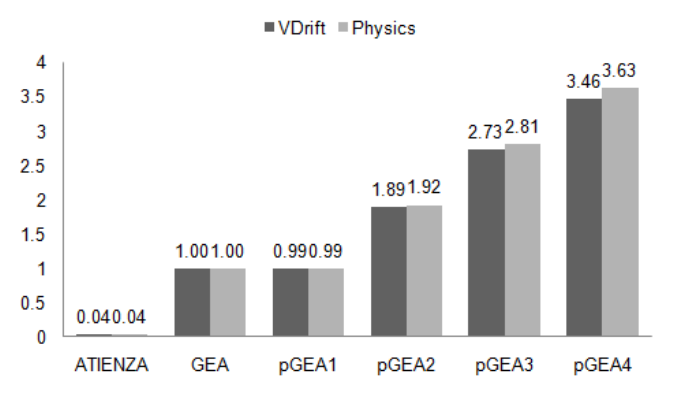}
\caption{Speed-ups of pGEA$_j$ (1, 2, 3 and 4 workers) optimization algorithms vs. GEA and Atienza's algorithm in VDrift and Physics3D.}
\label{fig:Performance}
\end{figure}

As Fig. \ref{fig:Performance} depicts, the slowest algorithm is Atienza, spending more than one week for the optimization of VDrift and more than 2 weeks for Physics. It is due to the manual evaluation of every candidate DMM. The fastest algorithm is pGEA$_4$ (4 workers), reaching speed-ups of 84.72$\times$ and 86.40$\times$ with respect to the Atienza's method, and 3.46$\times$ and 3.63$\times$ when compared to GEA for VDrift and Physics3D, respectively. Note that pGEA$_1$ does not improve the exploration speed when compared to GEA, reaching normalized exploration times of 0.99 and 0.99 for VDrift and Physics3D, respectively. It is because in this case there are two nodes, the first one is a master node running a GE algorithm and the other one is a worker performing the evaluation of the entire population, sequentially. Therefore, there is no parallelism in the evaluation process and, as a result, pGEA$_1$ has the highest execution time in both VDrift and Physics applications among all the five GE algorithms (both sequential and parallel) because of the DEVS intrinsic communication time.

From the experiments, we can state that our pGEA scales almost linearly with respect to the number of used workers. Obviously, the higher the number of workers, the larger the communication time, so the final gain in the parallelization is being decreased because of the communication process.

\subsection{On the quality of solutions}

In a second set of experiments, we compare the quality of the solutions found by our pGEA with four DMMs. The first one is the Kingsley DMM \cite{Wilson1995}. Although Kingsley is quite fast and extensively used in embedded operating systems (e.g., RTEMS~\cite{RTEMS} or Free BSD~\cite{Kozubik2001}), it can potentially present a considerable fragmentation due to its use of power-of-two segregated-fit lists. As a consequence, it uses more memory than other DMMs. The second one is the Lea DMM. Lea is one of the most complex general-purpose DMMs. It manages more than 128 lists for different block sizes. As a result, Lea presents less fragmentation than Kingsley. Thus, although Kinsgley is faster, Lea uses less amount of memory. The third DMM is a customized DMM generated using the methodology proposed by Atienza \emph{et al} \cite{Atienza2006}. The fourth DMM is the best DMM found by our sequential GEA.

In order to compare the quality of solutions between GEA and pGEA, we remain unchanged the genetic parameters shown in Table \ref{tab:Parameters} for GEA. Then, the fitness of the best DMM as well as the average GEA optimization time is set into pGEA$_4$ as the termination condition. Thus, pGEA$_4$ offers better DMMs than the sequential GEA approach in the same execution time.

Those comparisons of results are next described and analyzed distinguishing the two application benchmarks: VDrift and Physics3D.

\subsubsection{Method applied to VDrift}

\begin{figure*}[!t]
\centering
\subfloat[VDrift]{\label{fig:VDriftDMM}\includegraphics[width=0.49\textwidth]{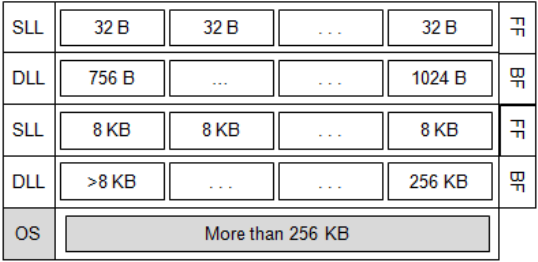}}
\hfill
\subfloat[Physics3D]{\label{fig:PhysicsDMM}\includegraphics[width=0.49\textwidth]{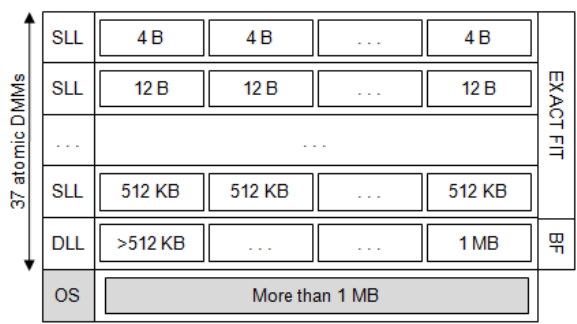}}
\caption{Custom DMM obtained using pGEA$_4$ for both VDrift and Physics3D applications.}
\label{fig:DMMResults}
\end{figure*}

The dynamic behavior of the VDrift case study shows that only a very limited range of data type sizes are used in it, namely 11 different allocation sizes are requested. In addition, most of these allocated sizes are relatively small (i.e., between 32 or 8192 Bytes) and only very few blocks are much bigger (e.g., 151 KBytes). Furthermore, we see that most of the data types interact with each other and are alive almost all the execution time of the application. Within this context, we apply our methodology using the order provided in Fig. \ref{fig:Framework}, optimizing performance, memory usage and energy consumed by the DMM. Fig. \ref{fig:VDriftDMM} shows our final solution offered by pGEA$_4$ that consists of a custom DMM with 4 separated ADMs for the relevant sizes in the application. The first ADM is used for the smallest allocation size requested in the application, that is, 32 bytes. The second ADM allows allocations of sizes between 756 bytes and 1024 bytes. Then, the third ADM is used for allocation requests of 8192 bytes. Finally, the fourth ADM is used for big allocation requests blocks (e.g., 151 or 265 KBytes). The ADM for the smallest size has its blocks in a singly-linked list because it does not need to coalesce or split because only one block size can be requested in it, as well as in the third ADM. The rest of the ADMs include doubly-linked lists of free blocks with headers that contain the size of each respective block and information about their current state (i.e., in use or free). These mechanisms efficiently support immediate coalescing and splitting inside these ADMs, which minimizes the total amount of memory used in the custom DMM designed with our methodology. Finally, the obtained DMM alternates between First Fit and Best Fit policies (FF and BF in Fig. \ref{fig:VDriftDMM}). Since we are managing unsortered lists, Best Fit is the best allocation policy in case the atomic DMM is designed for more than one block size.

Fig. \ref{fig:VDrift} shows that the values obtained by pGEA$_4$ using our parallel approach obtains significant improvements in all the three metrics. Regarding performance, the DMM obtained by pGEA$_4$ is 10.75\% faster than Kingsley (which is the fastest general-purpose DMM), 87.38\% faster than Lea, and 18.16\% and 10.33\% faster than Atienza and GEA, respectively. Our methodology achieves better results than previous classic optimizations because our custom DMM does not employ much time when looking for a free-block, due to the optimized size of the final DMM (for instance, 4 atomic DMMs vs. more than 128 regions in Lea).

With respect to memory usage, our custom pGEA$_4$ DMM saves 57.71\% more memory than Kingsley, 30.04\% more than Lea, 27.17\% more than Atienza, and 17.05\% more than GEA. This result is obtained because our pGEA$_4$ DMM is able to minimize the memory used by the system in two ways. First, because its design and behavior varies according to the different block sizes requested. Second, in ADMs where a range of block sizes requests are allowed, it uses immediate coalescing and splitting services to reduce both internal and external fragmentation, reducing the amount of memory used. In Kingsley and Lea DMMs, the coalescing/splitting mechanisms are applied, but an initial boundary memory is reserved and distributed among the different lists for sizes. In this case, since only a limited amount of sizes is used, some of the memory regions managed are underused. The Atienza's optimization method presented offered the worst results in this case, probably because this methodology needs and initial manual filtering of the search space, and some good potential solutions could be abandoned.

Concerning energy consumption, the pGEA$_4$ DMM reached the best value, 71.80\% better than Kingsley, and 87.48\%, 45.97\% and 12.44\% better than Lea, Atienza and GEA, respectively. Since the structure is more compact than in the other approaches, the pGEA$_4$ DMM needs less number of memory reads to find a free-block, and as a consequence, it consumes less energy.

\subsubsection{Method applied to Physics3D}

Fig. \ref{fig:PhysicsDMM} shows a graphic representation of the best DMM obtained using pGEA$_4$. First, as Fig. \ref{fig:PhysicsDMM} illustrates, it makes the decision to have 37 ADMs (more than 37 block sizes) to prevent internal fragmentation and as a consequence, to use less memory. This is done because the memory blocks requested by the Physics3D application vary greatly in size (to store bodies of different sizes) and if only one block size is used for all the different block sizes requested, the internal fragmentation would be large. Next, the algorithm chooses splitting or coalescing, so that every time a memory block with a bigger or smaller size than the current block is requested, the splitting and coalescing mechanisms are invoked. In addition, an exact fit to avoid as much as possible memory losses in internal fragmentation is selected. The last atomic DMM manages blocks varying from more than 512 KB to 1 MB. Bigger blocks are managed by the system (in gray background in Fig. \ref{fig:PhysicsDMM}). Finally, the pGEA$_4$ selects a header field to accommodate information about the size and status of each block to support splitting and coalescing mechanisms.

\begin{figure*}[!t]
\centering
\subfloat[VDrift]{\label{fig:VDrift}\includegraphics[width=0.49\textwidth]{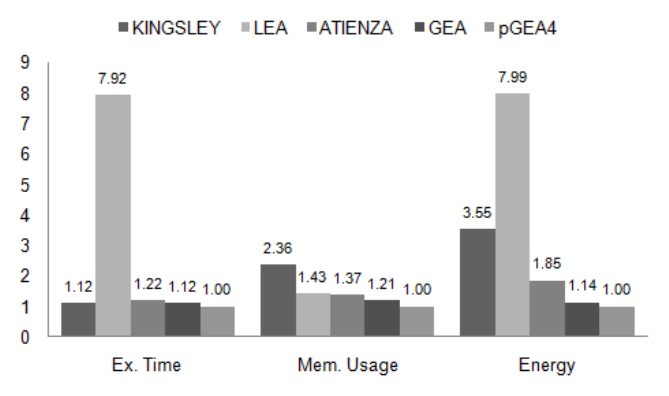}}
\subfloat[Physics3D]{\label{fig:Physics3D}\includegraphics[width=0.49\textwidth]{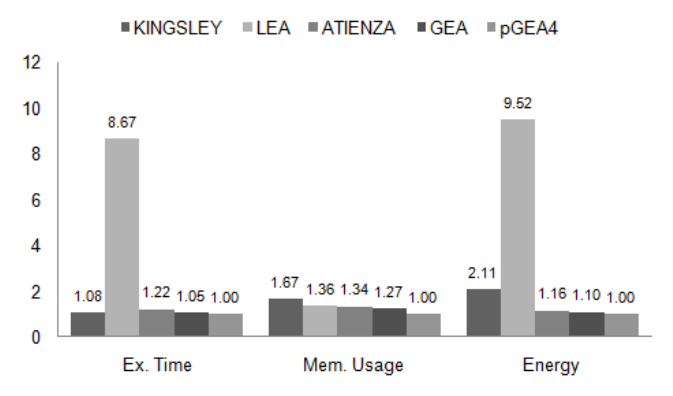}}
\caption{Results of execution time, memory usage and energy consumption of the set of DMMs formed by Kingsley, Lea, the custom DMM obtained by the methodology proposed in \cite{Atienza2006} (labeled as Atienza in the figure), and our custom DMMs based on grammatical evolution, both sequential and parallel approaches (labeled as GEA and pGEA4, respectively).}
\label{fig:NumericalResults}
\end{figure*}

As Fig. \ref{fig:Physics3D} shows, the fastest DMM is our pGEA$_4$ DMM, 7.55\%, 88.47\%, 17.86\% and 4.70\% faster than Kingsley, Lea, Atienza and GEA, respectively. It is because of the compact structure generated, which is optimal for the wide range of blocks managed by Physics3D in order to find free-blocks up to 88.47\% faster than Lea.

Regarding memory usage, our pGEA$_4$ DMM uses less memory (reduction of 40.11\%, 26.21\%, 25.16\% and 21.16\%) than the other four DMMs. This is due to the fact that our custom DMM manager does not have many fixed sized blocks to try with multiple accesses, and tries to coalesce and split as much as needed to efficiently used the existing memory, which is a better option in dynamic applications with large variations in requested sizes.

Furthermore, the results indicate that our pGEA$_4$ DMM achieves significantly better results for energy (a reduction of 52.50\%, 89.50\%, 14.08\% and 9.04\%), when compared to Kingsley, Lea, Atienza and GEA, because most of the dynamic accesses performed internally by the other four DMMs to its complex management structures are not required in the pGEA$_4$ DMM, which uses a simpler and optimized internal data structures for the target application.


\section{Conclusions and Future Work}
\label{sec:Conclusions}
Nowadays, high-performance embedded devices (e.g., PDAs, advanced mobile phones, portable video games stations, etc.) need to execute very dynamic embedded applications. These applications have been typically executed in general-purpose computers, so they are very complex for embedded systems and currently demand intensive dynamic memory requirements that must be heavily optimized (i.e., performance, memory usage and energy consumption) for an efficient mapping on latest low-power embedded devices. System-level exploration and refinement methodologies have started to be proposed to consistently perform that refinement. Within this context, the manual exploration and optimization of the DMM implementation is one of the most time-consuming and programming-intensive parts. In this paper we have presented a new system-level approach based on genetic programming to automatically characterize custom DMMs with an integrated profiling method. This approach largely simplifies the complex engineering process of designing and profiling several implementation candidates, allowing the developers to automatically cover a vast part of the dynamic memory management design space (e.g., different strategies of the heap, internal blocks of the allocators, etc.) without any programming and modelling effort. Particularly, we have proposed a new parallel grammatical {\it evolutionary algorithm (pGEA)}, which uses a master-worker scheme. The results obtained show that this parallel approach performs better, obtaining a speed-up of 86.40$\times$ with respect to other state-of-the-art approaches and a speed-up of 3.63$\times$ with respect to our previous sequential scheme. Furthermore, we have shown in our case studies that the pGEA is able to improve the quality of solutions in a 36.36\%, averaging the level of improvement in performance, memory usage and energy consumption obtained by four state-of-the-art approaches to this problem. The parallelization has been performed using the DEVS M\&S formalism. To this end, we have developed a DEVS master-worker model using a DEVS distributed library called DEVS/SOA. Future work includes the evaluation of several topologies as well as several synchronization policies.

\section*{Acknowledgments}
We would like to thank the reviewers for providing encouraging feedback and insightful comments that improved the content and the quality of the paper. This work has been supported by Spanish Government grants TIN2008-00508 and MEC Consolider Ingenio CSD00C-07-20811 of the Spanish Council of Science and Innovation. This research has been also supported by the Swiss NSF Research Project (Division II) Grant number 200021-127282.

\section*{References}

\bibliographystyle{elsarticle-num}
\bibliography{bibliography}

\begin{thebibliography}{10}
\expandafter\ifx\csname url\endcsname\relax
  \def\url#1{\texttt{#1}}\fi
\expandafter\ifx\csname urlprefix\endcsname\relax\def\urlprefix{URL }\fi
\expandafter\ifx\csname href\endcsname\relax
  \def\href#1#2{#2} \def\path#1{#1}\fi

\bibitem{RiscoMartin2008a}
J.~L. Risco-Martín, D.~Atienza, J.~I. Hidalgo, J.~Lanchares, A parallel
  evolutionary algorithm to optimize dynamic data types in embedded systems,
  Soft Comput. 12~(12) (2008) 1157--1167.
\newblock \href {http://dx.doi.org/10.1007/s00500-008-0295-y}
  {\path{doi:10.1007/s00500-008-0295-y}}.

\bibitem{Calder1994}
B.~Calder, D.~Grunwald, B.~Zorn, Quantifying behavioral differences between c
  and c++ programs, Journal of Programming Languages 2 (1995) 313--351.

\bibitem{Vijaykrishnan2003}
N.~Vijaykrishnan, M.~Kandemir, M.~J. Irwin, H.~S. Kim, W.~Ye, D.~Duarte,
  Evaluating integrated hardware-software optimizations using a unified energy
  estimation framework, IEEE Trans. Comput. 52~(1) (2003) 59--76.
\newblock \href {http://dx.doi.org/http://dx.doi.org/10.1109/TC.2003.1159754}
  {\path{doi:http://dx.doi.org/10.1109/TC.2003.1159754}}.

\bibitem{Wilson1995}
P.~R. Wilson, M.~S. Johnstone, M.~Neely, D.~Boles, Dynamic storage allocation:
  A survey and critical review, in: IWMM '95: Proceedings of the International
  Workshop on Memory Management, Springer-Verlag, London, UK, 1995, pp. 1--116.

\bibitem{Berger2001}
E.~D. Berger, B.~G. Zorn, K.~S. McKinley, Composing high-performance memory
  allocators, SIGPLAN Not. 36~(5) (2001) 114--124.
\newblock \href {http://dx.doi.org/http://doi.acm.org/10.1145/381694.378821}
  {\path{doi:http://doi.acm.org/10.1145/381694.378821}}.

\bibitem{Vo1996}
K.-P. Vo, Vmalloc: A general and efficient memory allocator, Software Practice
  and Experience 26 (1996) 1--18.

\bibitem{RTEMS}
{R}eal-{T}ime {O}perating {S}ystem for {M}ultiprocessor {S}ystems ({RTEMS}),
  http://www.rtems.com (2008).

\bibitem{Attardi1999}
G.~Attardi, T.~Flagella, P.~Iglio, A customisable memory management framework
  for c++, Software: Practice and Experience 28~(11) (1999) 1143--1184.

\bibitem{Atienza2006}
D.~Atienza, S.~Mamagkakis, F.~Poletti, J.~M. Mendias, F.~Catthoor, L.~Benini,
  D.~Soudris, Efficient system-level prototyping of power-aware dynamic memory
  managers for embedded systems, Integr. VLSI J. 39~(2) (2006) 113--130.
\newblock \href
  {http://dx.doi.org/http://dx.doi.org/10.1016/j.vlsi.2004.08.003}
  {\path{doi:http://dx.doi.org/10.1016/j.vlsi.2004.08.003}}.

\bibitem{Atienza2006a}
D.~Atienza, J.~M. Mendias, S.~Mamagkakis, D.~Soudris, F.~Catthoor, Systematic
  dynamic memory management design methodology for reduced memory footprint,
  ACM Trans. Des. Autom. Electron. Syst. 11~(2) (2006) 465--489.
\newblock \href {http://dx.doi.org/http://doi.acm.org/10.1145/1142155.1142165}
  {\path{doi:http://doi.acm.org/10.1145/1142155.1142165}}.

\bibitem{Mamagkakis2006}
S.~Mamagkakis, D.~Atienza, C.~Poucet, F.~Catthoor, D.~Soudris, Energy-efficient
  dynamic memory allocators at the middleware level of embedded systems, in:
  EMSOFT '06: Proceedings of the 6th ACM \& IEEE International conference on
  Embedded software, ACM, New York, NY, USA, 2006, pp. 215--222.
\newblock \href {http://dx.doi.org/http://doi.acm.org/10.1145/1176887.1176919}
  {\path{doi:http://doi.acm.org/10.1145/1176887.1176919}}.

\bibitem{RiscoMartin2009b}
J.~L. Risco-Mart\'{\i}n, D.~Atienza, R.~Gonzalo, J.~I. Hidalgo, Optimization of
  dynamic memory managers for embedded systems using grammatical evolution, in:
  GECCO '09: Proceedings of the 11th Annual conference on Genetic and
  evolutionary computation, ACM, New York, NY, USA, 2009, pp. 1609--1616.
\newblock \href {http://dx.doi.org/http://doi.acm.org/10.1145/1569901.1570116}
  {\path{doi:http://doi.acm.org/10.1145/1569901.1570116}}.

\bibitem{ONeill2003}
M.~O'Neill, C.~Ryan, Grammatical Evolution: Evolutionary Automatic Programming
  in an Arbitrary Language, Kluwer Academic Publishers, 2003.

\bibitem{CantuPaz1998}
E.~Cantú-Paz, A survey of parallel genetic algorithms, Calculateurs Paralleles,
  Reseaux et Systems Repartis 10~(2) (1998) 141--171.

\bibitem{Mittal2009a}
S.~Mittal, J.~L. Risco-Martín, B.~P. Zeigler, Devs/soa: A cross-platform
  framework for net-centric modeling and simulation in devs unified process,
  SIMULATION: Transactions of SCS 85~(7) (2009) 419--450.
\newblock \href {http://dx.doi.org/10.1177/0037549709340968}
  {\path{doi:10.1177/0037549709340968}}.

\bibitem{Lea2009}
D.~Lea, A memory allocator, http://g.oswego.edu/dl/html/malloc.html.

\bibitem{Stroustrup2000}
B.~Stroustrup, The C++ Programming Language, Addison-Wesley Longman Publishing
  Co., Inc., Boston, MA, USA, 2000.

\bibitem{Brabazon2006}
A.~Brabazon, M.~O'Neill, Biologically Inspired Algorithms for Financial
  Modelling, Springer, 2006.

\bibitem{Dempsey2007}
I.~Dempsey, M.~O'Neill, A.~Brabazon, Constant creation in grammatical
  evolution, Int. J. Innov. Comput. Appl. 1~(1) (2007) 23--38.
\newblock \href {http://dx.doi.org/http://dx.doi.org/10.1504/IJICA.2007.013399}
  {\path{doi:http://dx.doi.org/10.1504/IJICA.2007.013399}}.

\bibitem{ONeill2001}
M.~O'Neill, C.~Ryan, Grammatical evolution, IEEE Trans. Evolutionary
  Computation 5~(4) (2001) 349--358.

\bibitem{Brabazon2008}
A.~Brabazon, M.~O'Neill, I.~Dempsey, An introduction to evolutionary
  computation in finance, Computational Intelligence Magazine, IEEE 3~(4)
  (2008) 42--55.
\newblock \href {http://dx.doi.org/10.1109/MCI.2008.929841}
  {\path{doi:10.1109/MCI.2008.929841}}.

\bibitem{Poli2008}
R.~Poli, W.~B. Langdon, N.~F. McPhee, A field guide to genetic programming,
  Published via http://lulu.com and freely available at
  http://www.gp-field-guide.org.uk., 2008.

\bibitem{Mamidipaka2004}
M.~Mamidipaka, N.~Dutt, e{CACTI}: {A}n enhanced power estimation model for
  on-chip caches, Tech. Rep. TR-04-28, CECS, UC Irvine (2004).

\bibitem{Sipser2005}
M.~Sipser, Introduction to the Theory of Computation, Course Technology, 2005.

\bibitem{deb2001}
K.~Deb, Multiobjective Optimization using Evolutionary Algorithms, John Wiley
  and Son Ltd., 2001.

\bibitem{Fernandez2003}
F.~Fernández, M.~Tomassini, L.~Vanneschi, An empirical study of multipopulation
  genetic programming, Genetic Programming and Evolvable Machines 4~(1) (2003)
  21--51.
\newblock \href {http://dx.doi.org/http://dx.doi.org/10.1023/A:1021873026259}
  {\path{doi:http://dx.doi.org/10.1023/A:1021873026259}}.

\bibitem{Fernandez2005}
F.~Fernandez, G.~Spezzano, M.~Tomassini, L.~Vanneschi, Parallel Genetic
  Programming, Parallel and Distributed Computing, Wiley-Interscience, Hoboken,
  New Jersey, USA, 2005, Ch.~6, pp. 127--153.

\bibitem{Hidalgo2007}
J.~I. Hidalgo, J.~Lanchares, F.~Fernández~de Vega, D.~Lombraña, Is the island
  model fault tolerant?, in: GECCO '07: Proceedings of the 2007 GECCO
  conference companion on Genetic and evolutionary computation, ACM, New York,
  NY, USA, 2007, pp. 2737--2744.
\newblock \href {http://dx.doi.org/http://doi.acm.org/10.1145/1274000.1274085}
  {\path{doi:http://doi.acm.org/10.1145/1274000.1274085}}.

\bibitem{CantuPaz1998b}
E.~Cantú-Paz, Designing efficient master-slave parallel genetic algorithms, in:
  Genetic Programming: Proc. of the Third Annual Corference, 1998, pp.
  455--460.

\bibitem{Zeigler2000}
B.~P. Zeigler, T.~Kim, H.~Praehofer, Theory of Modeling and Simulation:
  Integrating Discrete Event and Continuous Complex Dynamic Systems, Academic
  Press, 2000.

\bibitem{RiscoWeb2009}
J.~L. Risco-Martin, D.~Atienza, D{DT} evaluation repository using java
  evolutionary computation library (jeco), Available at:
  http://www.dacya.ucm.es/jlrisco (2009).

\bibitem{Gropp1999}
W.~Gropp, R.~Thakur, E.~Lusk, Using MPI-2: Advanced Features of the Message
  Passing Interface, MIT Press, Cambridge, MA, USA, 1999.

\bibitem{Lima2009}
J.~V. Lima, N.~Maillard, Online mapping of mpi-2 dynamic tasks to processes and
  threads, International Journal of High Performance Systems Architecture 2~(2)
  (2009) 81--89.
\newblock \href {http://dx.doi.org/10.1504/IJHPSA.2009.032025}
  {\path{doi:10.1504/IJHPSA.2009.032025}}.

\bibitem{HLA2000}
IEEE, Standard for modeling and simulation ({M\&S}) high level architecture
  ({HLA}), Tech. Rep. 1516, IEEE (2000).

\bibitem{VDrift}
{VD}rift racing simulator, http://vdrift.net (2008).

\bibitem{Kharevych2002}
L.~Kharevych, R.~Khan, 3{D} physics engine for elastic and deformable bodies,
  University of Maryland, College Park (2002).

\bibitem{ONeill2008}
M.~O'Neill, E.~Hemberg, C.~Gilligan, E.~Bartley, J.~McDermott, A.~Brabazon,
  G{EVA} - grammatical evolution in java, SIGEVOlution 3~(2) (2008) 17--22.
\newblock \href {http://dx.doi.org/http://doi.acm.org/10.1145/1527063.1527066}
  {\path{doi:http://doi.acm.org/10.1145/1527063.1527066}}.

\bibitem{Kozubik2001}
J.~Kozubik, Free{BSD} and solid state devices, Available at:
  http://www.freebsd.org/doc/en/articles/solid-state/index.html (2001).

\end{thebibliography}

\end{document}